\documentclass[a4paper,twoside]{article}

\usepackage{epsfig}
\usepackage{subcaption}
\usepackage{calc}
\usepackage{amssymb}
\usepackage{amstext}
\usepackage{amsmath}
\usepackage{amsthm}
\usepackage{multicol}
\usepackage{pslatex}
\usepackage{apalike}
\usepackage{xcolor}
\usepackage{colortbl}
\usepackage[pagebackref=false,breaklinks=true,letterpaper=true,colorlinks=true,citecolor=black,bookmarks=false]{hyperref}

\usepackage{SCITEPRESS}     

\begin{document}

\title{FisheyeSuperPoint: Keypoint Detection and Description Network for Fisheye Images}

\author{\authorname{Anna Konrad\sup{1, 2}, Ciarán Eising\sup{3}, Ganesh Sistu\sup{4}, John McDonald\sup{5}, Rudi Villing\sup{2}, Senthil Yogamani\sup{4}}
\affiliation{\sup{1}Hamilton Institute, Maynooth University, Ireland}
\affiliation{\sup{2}Department of Electronic Engineering, Maynooth University, Ireland}
\affiliation{\sup{3}Department of Electronic \& Computer Engineering, University of Limerick, Ireland}
\affiliation{\sup{4}Valeo Vision Systems, Galway, Ireland}
\affiliation{\sup{5}Department of Computer Science, Maynooth University, Ireland}
\email{anna.konrad.2020@mumail.ie, ciaran.eising@ul.ie, ganesh.sistu@valeo.com}
}


\abstract{Keypoint detection and description is a commonly used building block in computer vision systems particularly for robotics and autonomous driving.
However, the majority of techniques to date have focused on standard cameras with little consideration given to fisheye cameras which are commonly used in urban driving and automated parking. In this paper, we propose a novel training and evaluation pipeline for fisheye images. We make use of SuperPoint as our baseline which is a self-supervised keypoint detector and descriptor that has achieved state-of-the-art results on homography estimation. 
We introduce a fisheye adaptation pipeline to enable training on undistorted fisheye images. We evaluate the performance on the HPatches benchmark, and, by introducing a fisheye based evaluation method for detection repeatability and descriptor matching correctness, on the Oxford RobotCar dataset.}

\onecolumn \maketitle \normalsize \setcounter{footnote}{0} \vfill

\section{\uppercase{Introduction}}
\label{sec:introduction}

Keypoint detection and description is a fundamental step in computer vision for image registration \cite{ma2021imagematching}. 
It has a wide range of applications including 3D reconstruction, object tracking, video stabilization and SLAM. Approaches designed to be invariant to changes in scale, illumination, perspective, etc. have been extensively studied in the computer vision literature. However, despite their prevalence in automotive and robotic systems, few approaches have explicitly considered fisheye images that pose an additional challenge of spatially variant distortion. Thus a patch in the centre of an image looks different compared to a region in the periphery of the image where the radial distortion is much higher. Fisheye cameras are a fundamental sensor in autonomous driving necessary to cover the near field around the vehicle \cite{yogamani2019woodscape}. Four fisheye cameras on each side of the vehicle can cover the entire $360^{\circ}$ field of view. 

A common approach to use fisheye images for tasks in computer vision is to first rectify the image and then apply algorithms suited for standard images \cite{lo2018fisheyestiching}, \cite{esparza20143dreconstruction}. A principal drawback of such methods is that the field of view is reduced and resampling artifacts can be introduced into the periphery of the image.
An alternative approach, as demonstrated in \cite{kumar2020unrectdepthnet}, is to work directly in the fisheye image space and thereby avoid such issues. Recently, there has been progress with such approaches for various visual perception tasks such as dense matching \cite{haene2014densematching}, object detection \cite{rashed2021generalized}, depth estimation \cite{kumar2018near}, re-localisation \cite{tripathi2020trained}, soiling detection \cite{uricar2019desoiling} and people detection \cite{duan2020rapid}.

Feature detectors and descriptors can describe corners (also called interest points or keypoints), edges or morphological region features. Traditionally, feature detection and description has been done with handcrafted algorithms \cite{ma2021imagematching}. Some of the most well-known algorithms include Harris \cite{harris1998}, FAST \cite{rosten2006fast} and SIFT \cite{lowe2004}. 
Thorough reviews of traditional and modern techniques have been conducted in various surveys \cite{ma2021imagematching}, \cite{mikolajczyk2005}, \cite{li2015survey}.

Recently, several CNN based feature correspondence techniques have been explored which outperform classical features. For example, a universal correspondence network in \cite{choy2016universal} demonstrates state-of-the-art results on various datasets by making use of a spatial transformer to normalise for affine transformations. This is an example of feature correspondence learning independent of the application in which it is used. It is an open problem to learn feature correspondence which is optimal for the later stages in the perception pipeline e.g. bundle adjustment. For instance, end-to-end learning of feature matching could possibly learn diversity and distribution rather than focusing solely on measures such as distinctiveness and repeatability.
This is particularly useful for training on fisheye images which have a domain gap compared to regular images. In addition, learning based detectors permit the encoder backend to be merged into a multi-task model. Such multi-task models share the same encoder, which can improve performance and reduce latency for all tasks \cite{leang2020dynamic}, \cite{kumar2021omnidet}.

SuperPoint is a self-supervised CNN framework for keypoint detection and description \cite{detone2018superpoint}. It consists of one encoder and two different decoders for the detection and the description output. It is pretrained as an corner detector via a synthetically generated dataset containing basic shapes like rectangles, lines, stars, etc. The resulting corner detector shows inconsistent keypoint detections on the same scene for varying camera viewpoints. 
To improve on the detection consistency for varying camera viewpoints, homographic adaptation is used to generate a superset of keypoints from random homographic warpings on the MS-COCO training dataset \cite{lin2014mscoco}. The network is then trained on those keypoints and the whole process is repeated for several iterations.
Recently, it has achieved state-of-the-art results in several benchmarks in combination with the SuperGlue matcher \cite{sarlin2020superglue}.

We adapt the SuperPoint feature detector and descriptor so it can be trained and evaluated on fisheye images. The contributions of this paper include:
\begin{itemize}
    \item Random fisheye warping and unwarping in place of homographic transforms.
    \item Implementation of fisheye adaption for self-supervised training of the SuperPoint network.
    \item Evaluation of the repeatability of detectors and comparison of their performance to FisheyeSuperPoint on fisheye and standard images.
    \item Evaluation of the matching correctness of descriptors and comparison of their performance to FisheyeSuperPoint on fisheye \& standard images.
\end{itemize}

\section{\uppercase{Method}}

We propose FisheyeSuperPoint, a SuperPoint network that has been trained self-supervised on fisheye images. To enable this, the homographic adaptation step in the training pipeline has been exchanged with our fisheye adaptation process to cope with the nonlinear mapping of fisheye images. 

\subsection{Datasets Used}

\begin{figure}[t]
\centerline{\includegraphics[width=0.45\textwidth]{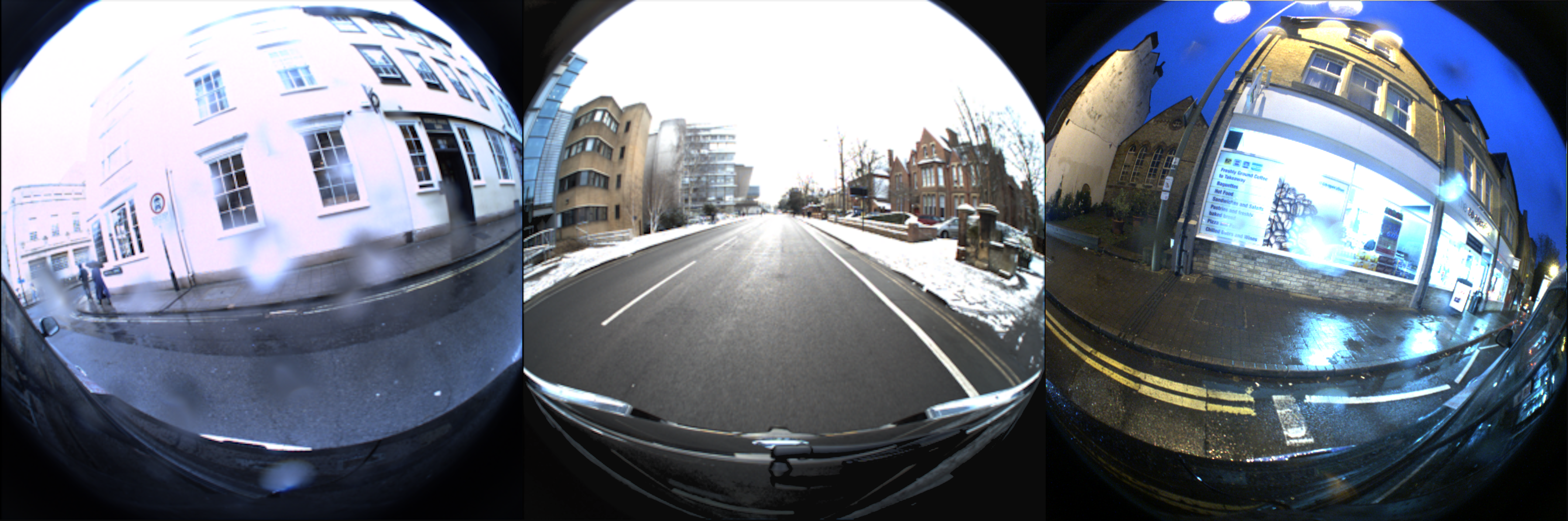}}
\caption{Example imagery of the Oxford RobotCar Dataset \cite{maddern2017robotcar}, showing different weather and lighting conditions.}
\label{fig_1}
\end{figure}

The data used for training of the FisheyeSuperPoint network is a subset of the Oxford RobotCar Dataset \cite{maddern2017robotcar} (RobotCar). The RobotCar dataset contains fisheye images, stereo images, LIDAR sensor readings and GPS from a vehicle across various seasons and weather conditions. Some examples from the fisheye images are shown in Figure \ref{fig_1}. The data is made available in subsets for each drive. The subsets contain data from three different fisheye cameras which were mounted on the left, right and rear of the vehicle. To generate a representative subset of fisheye images, image sequences from each of seven different weather conditions (sun, clouds, overcast, rain, snow, night, dusk) were used with 840k images in total. To reduce the total number of images and duplicates in the resulting dataset, we sampled every 10th frame from the sequences. 

The resulting training dataset contained 84k 1024x1024 fisheye images. During training, the images were downsampled to 256x256 images to match the resizing of MS-COCO \cite{lin2014mscoco} that was used for the training of SuperPoint \cite{detone2018superpoint}.

\subsection{Fisheye Warping and Unwarping}
\begin{figure*}[t]
\centerline{\includegraphics[width=0.95\textwidth]{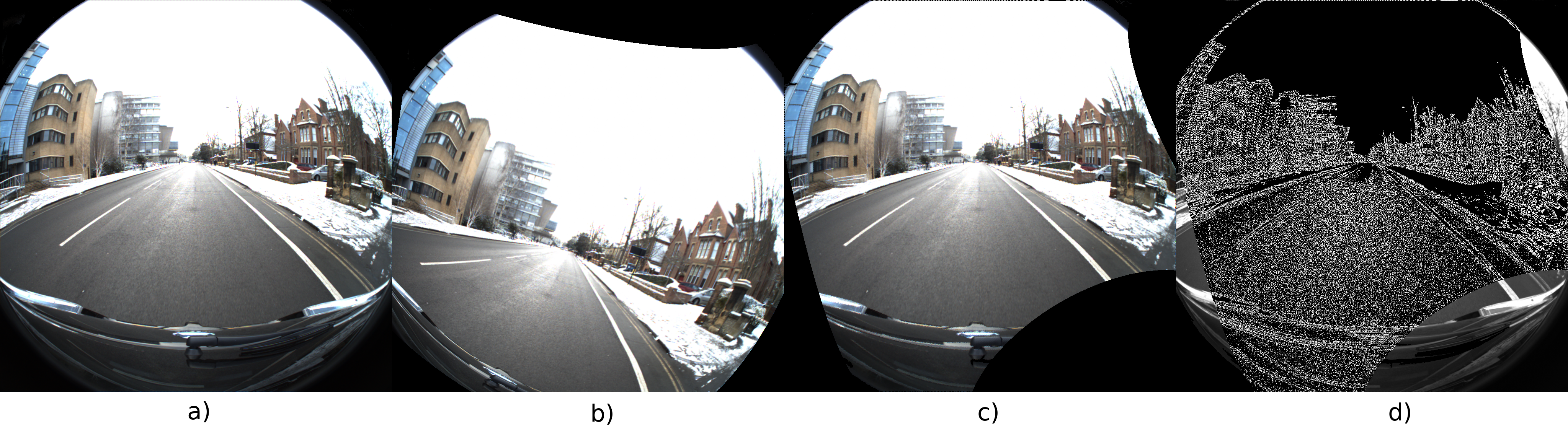}}
\caption{Results of the Fisheye Warping. a) the original image, b) the warped image, c) the unwarped image and d) the difference between the original and unwarped image. c) is the result of applying the inverse of the fisheye warping to b). The structural differences in d) are regions which were outside of the warped image b). The remaining differences in d) are due to the bilinear interpolation used for warping and unwarping the image.}
\label{fig_3}
\end{figure*}

In order to train FisheyeSuperPoint on fisheye images, a substitute for the homographic warping in SuperPoint is needed. The pinhole camera model is the standard projection function for much research in computer vision, as in SuperPoint \cite{detone2018superpoint}. The pinhole projection function is given as $\mathbf{p} = \left(\frac{fX}{Z},\frac{fY}{Z}\right)^\top$, where $\mathbf{X} = (X,Y,Z)^\top$ is a point in the camera coordinate system, and $f$ is the nominal \textit{focal length} of the pinhole camera.

Fisheye functions provide a nonlinear mapping from the camera coordinate system (e.g. Figure \ref{fig_3}). We can define a mapping from $\mathbb{R}^3$ to the fisheye image as
$$\pi: \mathbb{R}^3 \rightarrow I^2$$

A true inverse is naturally not possible, as all depth information is lost in the formation of the image. However, we can define an unprojection mapping from the fisheye image domain to the unit central projective sphere:
$$\pi^{-1}: {I}^2 \rightarrow {S}^2$$

Unfortunately, the Oxford RobotCar Dataset does not provide details of the fisheye model they use, nor its parameters. However, they provide a look-up-table that can map from a distorted to an undistorted image. We use this look-up-table and fit the fourth order polynomial model $p(\theta)$ described below.

In principle, it does not matter exactly which fisheye mapping function is used, as long as it provides a reasonably accurate model of the image transformation. In our case, we use a radial polynomial function for $\pi$, as per \cite{yogamani2019woodscape}:
\begin{align} \label{eqn:poly_fisheye}
    \pi(\mathbf{X}) &= \frac{p(\theta)}{d}\left[
    \begin{matrix} 
        X \\
        Y
    \end{matrix}
    \right] \nonumber, \quad d = \sqrt{X^2 + Y^2} \nonumber \\
    p(\theta) &= a_1 \theta + a_2 \theta^2 + \ldots + a_n \theta^n  \nonumber \\
    \theta &= \arccos \left(\frac{Z}{\sqrt{X^2 + Y^2 + Z^2}}\right)
\end{align}
where $p(\theta)$ is a polynomial of order $n$, with $n=4$ typically sufficient. 

In SuperPoint \cite{detone2018superpoint}, random homographies are used to simulate multiple camera viewpoints. In order to train a SuperPoint network with fisheye images, the homographic warping needs to be replaced with an equivalent transform that is applicable to fisheye imagery. Using the fisheye functions ($\pi$ and $\pi^{-1}$) described above, we can consider the following steps to generate a new fisheye warped image:
\begin{enumerate}
   \item Each point is projected from the image $I^2$ to a unit sphere $S^2$ using $\pi^{-1}$
   \item A new virtual camera position is selected by a random rotation $\mathbf{R}$ and translation $\mathbf{t}$ (six degrees of freedom Euclidean transform)
   \item Each point on the unit sphere is reprojected to this new, virtual camera position, by applying $\pi(\mathbf{R}\mathbf{X} + \mathbf{t})$
\end{enumerate}

This results in a mapping from $I^2 \rightarrow I'^2$, where $I'^2$ represents the new image with the random Euclidean transform applied. We call this mapping fisheye warping $\mathcal{F}$ and fisheye unwarping $\mathcal{F}^{-1}$:
\begin{align} \label{eqn:fisheye_warping}
    \mathcal{F}(I^2) = \pi(\mathbf{R}\pi^{-1}(I^2)+\mathbf{t})
\end{align}

\begin{figure*}[t!]
\centerline{\includegraphics[width=0.95\textwidth]{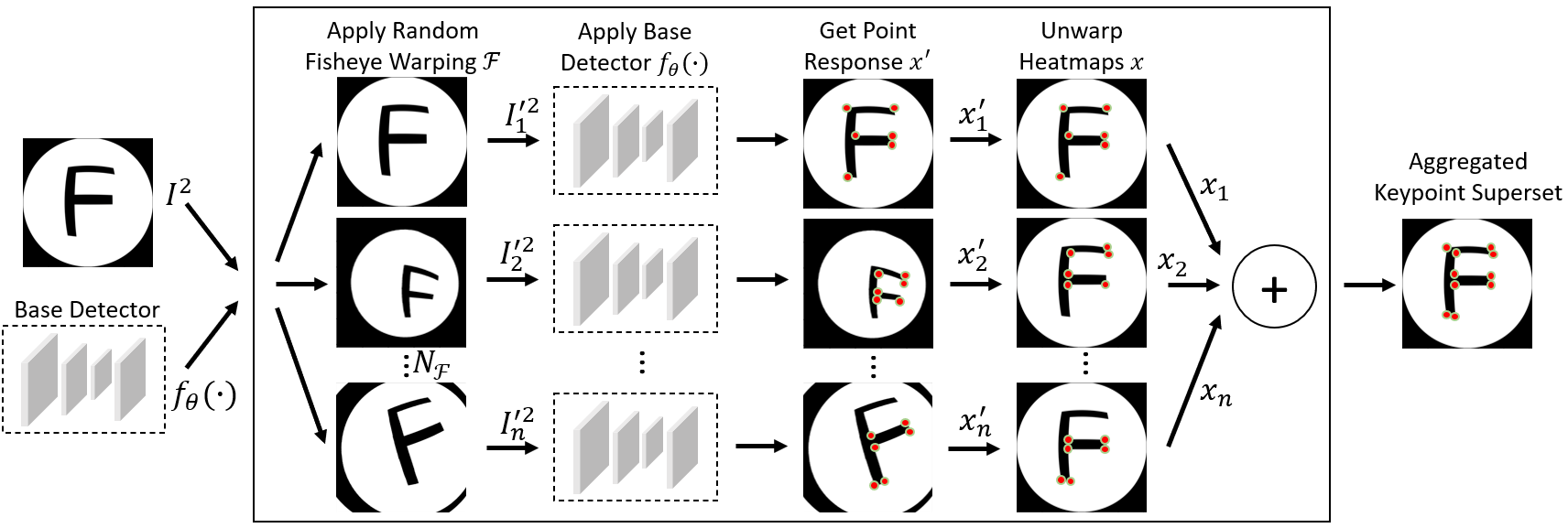}}
\caption{Self-supervised fisheye Keypoint Detection and Description training framework adapted from SuperPoint framework \cite{detone2018superpoint}. Random fisheye warpings are applied to a single image. Point responses are received from the base detector when applying it to the warped images. The point responses are then unwarped and accumulated to get an aggregated keypoint superset of the original image.
}
\label{fig_2}
\end{figure*}

In practice, to avoid sparsity in the new image, each pixel on the warped image is inverse transformed to the corresponding sub-pixel location on the original image and sampled using bilinear interpolation. Additionally, as the fisheye unprojection function $\pi^{-1}$ is computationally costly due to the requirement for a polynomial root solver, a set of 2000 look-up-tables is pre-computed for the image warpings. For each look-up-table, we sample the rotation and translation along each axis uniformly at random from the interval $U_R \in [-30^{\circ}, 30^{\circ}]$ and $U_t \in [-0.3, 0.3]$ relative to the unit sphere $S^2$, respectively.
In order to invert this warping, as required for unwarping the detected point responses, the inverse of the steps 1 - 3 above are followed. An example of the resulting images is shown in Figure \ref{fig_3}. The original image a) is warped to b) and unwarped to c) using the pre-computed look-up-tables. The difference d) between the original a) and unwarped image c) shows no structural differences apart from the regions which were outside of the warped image b) and therefore remain black in the unwarped image. The other differences are due to the bilinear interpolation used for warping and unwarping the image. 

\subsection{Fisheye Adaptation}

The fisheye warping is incorporated into the SuperPoint training pipeline as shown in Figure \ref{fig_2}. The aim of the adaptation process is to provide consistent keypoint responses on the same scene under varying camera viewpoints.
A single fisheye image $I^2$ and a base keypoint detector $f_{\theta}$ are input to the fisheye adaptation process. Within that process, a random fisheye warping $\mathcal{F}$ is sampled and applied to the image with bilinear interpolation resulting in a warped fisheye image $I'^2$. The base detector is used to detect point responses $x'$, which are subsequently unwarped back to the original image space by applying the inverse fisheye warping $\mathcal{F}^{-1}$.

The adaptation process is repeated for $N_{\mathcal{F}} = 100$ random fisheye warpings and the resulting point responses are accumulated. The point responses from different fisheye warpings are expected to differ initially. By accumulating the point responses into a superset of detected keypoints, a new ground-truth for keypoints in the original image is generated. Subsequently, the base detector is trained on this superset to generate a superior, more consistent keypoint detector. The full process can be repeated iteratively in order to enhance performance and consistency of the resulting model. The descriptor decoder is learned in a semi-dense manner in the last iteration of the model enhancement, as described in \cite{detone2018superpoint}.

\section{\uppercase{Results}}

FisheyeSuperPoint is developed on top of a trainable tensorflow implementation \cite{pautrat2021superpoint} based on SuperPoint \cite{detone2018superpoint}. To train FisheyeSuperPoint, we use a magic-point network \cite{pautrat2021superpoint} trained on MS-COCO as a base detector, where we apply fisheye adaptation to the RobotCar dataset and train FisheyeSuperPoint with 600,000 iterations. To train SuperPoint, we use the same pretrained network, where we apply homographic adaptation to the MS-COCO dataset and train SuperPoint with 600,000 iterations. 

The training including two fisheye/homographic adaptation iterations is executed on two Nvidia GTX 1080Ti GPUs and takes approximately one week to complete. The duration of the training with fisheye adaptation is similar to the training with homographic adaptation in SuperPoint. While the fisheye adaptation is a more complex procedure, the use of pre-calculated look-up-tables for the fisheye warping means it is computationally efficient. Note that the same testing data is used for all models in the experiments.

\subsection{Benchmark Setup}

\textbf{HPatches:} The performance of FisheyeSuperPoint and SuperPoint in comparison to traditional corner detection techniques is evaluated by the repeatability of detections and the homography estimation correctness on the HPatches benchmark \cite{balntas2017hpatches}. It contains multiple images of planar objects from varying camera viewpoints or with different illuminations. As the ground truth homography is provided for each corresponding image pair, detections on an image pair can be warped and their consistency can be compared. The detection repeatability of FisheyeSuperPoint and SuperPoint is compared to the detection algorithms FAST \cite{rosten2006fast}, Harris \cite{harris1998} and Shi \cite{shi1994}. The homography estimation correctness of FisheyeSuperPoint and SuperPoint is compared to SIFT \cite{lowe2004} and ORB \cite{rublee2011orb}. The evaluation methodology is the same as described in \cite{detone2018superpoint}, where the homography is estimated with OpenCV based on nearest neighbour matching on the descriptors.

The estimated homography is compared to the ground truth homography of HPatches by transforming four corner points $c_j$ of the image with both homographies, resulting in $c_j'$ and $\hat{c}_j'$ with $j = 4$. As per \cite{detone2018superpoint}, the homography estimation correctness is then calculated based on the distance between the corner points by $$H_c = \frac{1}{4}\sum_{j = 1}^4 ||c_{j}' - \hat{c}_{j}'|| \leq \epsilon$$ and averaged over all $n = 295$ test images.

\begin{table*}[tbp]
\caption{Detector Repeatability on HPatches and RobotCar}
\begin{center}
\begin{tabular}{c c c c c}
\hline
Algorithm & \multicolumn{2}{c}{Illumination Changes} & \multicolumn{2}{c}{Viewpoint Changes}\\
& NMS=4 & NMS=8 & NMS=4 & NMS=8 \\
\hline
\multicolumn{5}{c}{HPatches}\\
\hline
\textit{FisheyeSuperPoint}  & \textbf{0.664} & \textbf{0.631}& 0.678        & \textbf{0.626} \\
\textit{SuperPoint \cite{detone2018superpoint}} & 0.663         & 0.622         & 0.672         & 0.610\\
\textit{FAST \cite{rosten2006fast}} & 0.576 & 0.493         & 0.598         & 0.492\\
\textit{Harris \cite{harris1998}} & 0.630   & 0.590         & \textbf{0.725} & 0.612  \\
\textit{Shi \cite{shi1994}}  & 0.584        & 0.515         & 0.613         & 0.523\\
\hline
\multicolumn{5}{c}{RobotCar}\\
\hline
\textit{FisheyeSuperPoint}                      & 0.896          & \textbf{0.876} & 0.768       & \textbf{0.716} \\
\textit{SuperPoint \cite{detone2018superpoint}} & \textbf{0.897} & 0.869         & 0.754         & 0.708 \\
\textit{FAST \cite{rosten2006fast}}             & 0.837          & 0.751         & 0.724         & 0.569 \\
\textit{Harris \cite{harris1998}}               & 0.876          & 0.841         & \textbf{0.827} & 0.693 \\
\textit{Shi \cite{shi1994}}                     & 0.831          & 0.769         & 0.709         & 0.604\\
\hline
\end{tabular}
\label{tab1}
\end{center}
\end{table*}

\begin{table*}[tbp]
\caption{Homography correctness on HPatches, descriptor matching correctness and RMSE on RobotCar.}
\begin{center}
\begin{tabular}{c c c c}
\hline
Algorithm & HPatches $H_c$ & RobotCar $M_c$ & RobotCar $RMSE$\\
\hline
\textit{FisheyeSuperPoint}          & 0.712             & \textbf{0.862} & 38.4\\
\textit{SuperPoint \cite{detone2018superpoint}} & 0.668 & 0.859         & \textbf{36.3}\\
\textit{SIFT \cite{lowe2004}}       & \textbf{0.766}    & 0.663         & 120.7\\
\textit{ORB \cite{rublee2011orb}}    & 0.414            & 0.463         & 136.9\\
\hline

\end{tabular}
\label{tab2}
\end{center}
\end{table*}

\textbf{Fisheye Oxford RobotCar:}
In order to assess the performance of the trained networks on fisheye images, we evaluate keypoint detection repeatability and matching correctness using a fisheye test set. Unfortunately, there currently is no fisheye image equivalent to the HPatches benchmark available that contains precise ground truth viewpoint relations. Therefore, to evaluate performance on fisheye images, we generate an artificial dataset based on a test set of RobotCar \cite{maddern2017robotcar}. The test set was generated from approximately 51k images across five different weather conditions and image sequences that had not been used in the training set for FisheyeSuperPoint. To increase diversity in the test set, one out of every 172 frames was sampled resulting in a base test set of 300 images.

From the base test set, 300 illumination change test images are created by applying gamma correction, with gamma for each image drawn randomly from a uniform distribution $\gamma \in [0.1, 2]$. Independently, each image of the base test set is warped to create a viewpoint change test image using a fisheye warping $\mathcal{F}$ from a set of 300 random fisheye warpings not previously used for training. 

By applying the same keypoint detector to corresponding images we generate two different point responses:
$$ x = f_{\theta}(I^2), \quad\quad  x' = f_{\theta}(I'^2)$$
In the case of viewpoint changes, point responses that lie outside the overlapping region of both images are filtered by applying boolean masks. The masks are generated by warping all-ones matrices of image size with $\mathcal{F}$ and $\mathcal{F}^{-1}$ for $x'$ and $x$, respectively.
We apply $\mathcal{F}(x)$ to warp the point response $x$ into the image space of $x'$ and calculate the detector repeatability as described in \cite{detone2018superpoint}.

In order to evaluate the descriptor matching correctness $M_c$ on RobotCar, nearest neighbour matching is performed on the descriptors, resulting in a set of matches $M$. The keypoints $x_m$ of the resulting matches in $I^2$ are warped using $\mathcal{F}(x_m)$ and the euclidean distance $d()$ to their corresponding keypoint, $x_m'$, is calculated. The inliers $G$ are calculated as $G = \{x_m, x_m' \in \mathbb{R}^{2} : d(x_m, x_m') < \epsilon\}$. The threshold with $\epsilon = 3px$ is set to the same distance as for the homography estimation correctness. The matching correctness $M_c$ is calculated as the ratio of inliers to matches $M_c = \frac{|G|}{|M|}$, where $|G|$ denotes the number of elements in the set $G$. We average the descriptor matching correctness $M_c$ over all $n = 300$ test images for one given model. 

In addition, we report the root-mean-square error (RMSE) of the distance in all $i = n \times k$ matches by 
\begin{align} \label{eqn:rmse}
    RMSE = \sqrt{\frac{d_1^2 + d_2^2 + \dots + d_i^2}{i}}
\end{align}
for one given model. The number of keypoint detections $k = 300$ and the number of test images $n = 300$ is kept constant for all experiments.

\subsection{Comparison}

A comparison of the detection repeatability on FisheyeSuperPoint, SuperPoint, as well as FAST \cite{rosten2006fast}, Harris \cite{harris1998} and Shi \cite{shi1994} is shown in Table \ref{tab1}. The default OpenCV implementation is used for FAST, Harris and Shi. We apply Non-Maximum Suppression (NMS) on a square mask with size of $4px, 8px$ to the keypoint detections. The number of detected points $k = 300$ and correct distance threshold $ \epsilon = 3$ stays constant. Images in HPatches are resized to $240 \times 320$ and RobotCar images are resized to $256 \times 256$. 

FisheyeSuperPoint outperforms the other detectors for viewpoint and illumination changes in RobotCar when a high NMS is applied. While the traditional Harris detector outperforms FisheyeSuperPoint with a NMS $= 8$ on viewpoint changes, it ranks second and is superior to SuperPoint, FAST and Shi.

\begin{figure*}[t]
\centerline{\includegraphics[width=\textwidth]{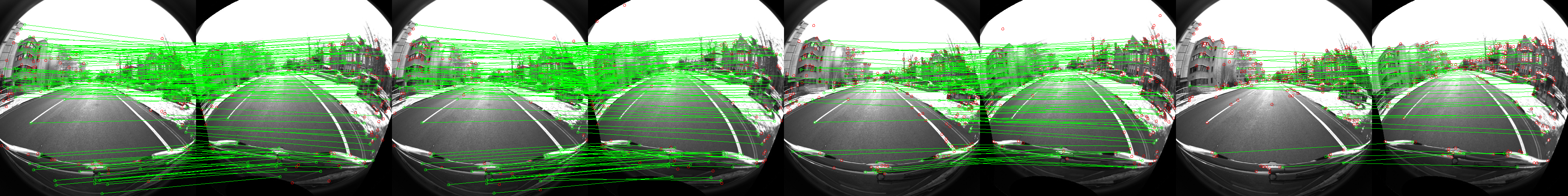}}
\centerline{\includegraphics[width=\textwidth]{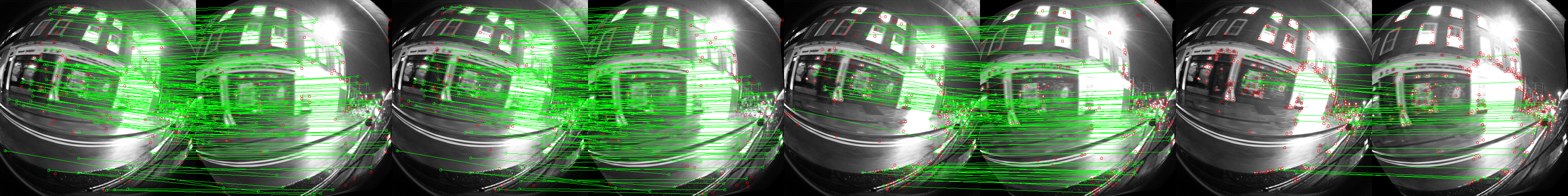}}
\centerline{\includegraphics[width=\textwidth]{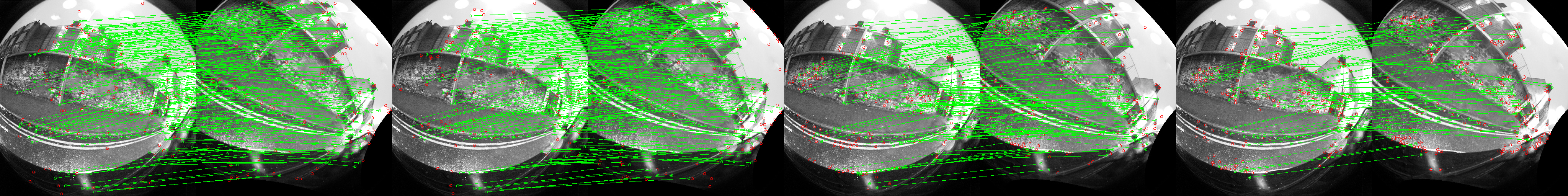}}
\centerline{\includegraphics[width=\textwidth]{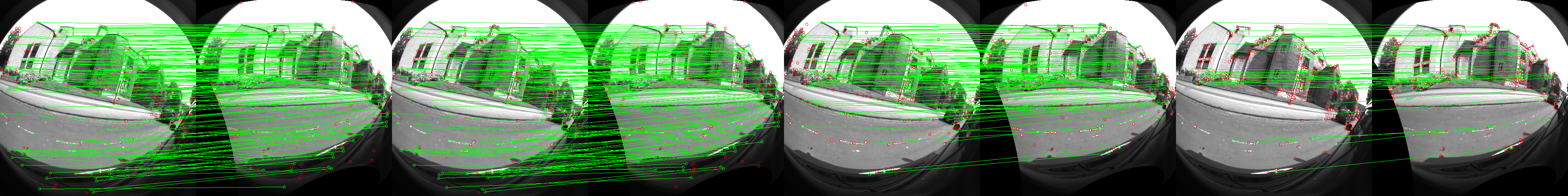}}
\centerline{\includegraphics[width=\textwidth]{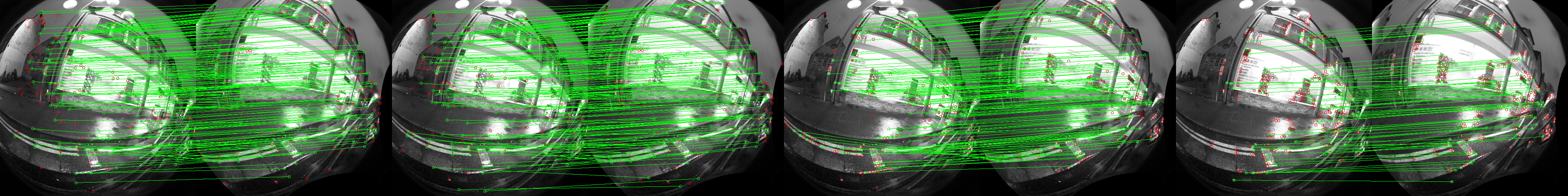}}
\centerline{\includegraphics[width=\textwidth]{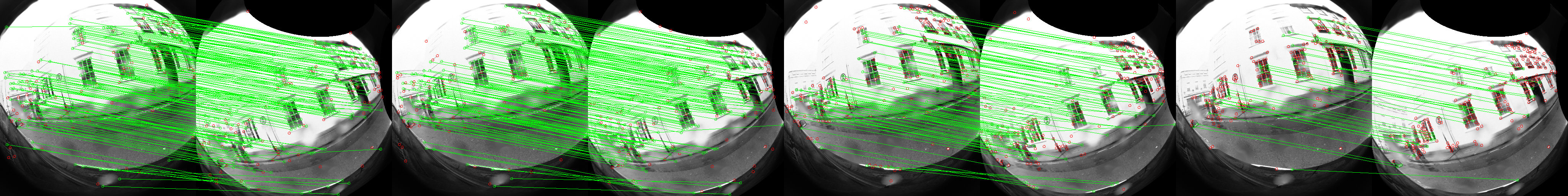}}
\caption{Qualitative results of feature matching on RobotCar images with $k = 300$ detected points. Nearest neighbour matches with a distance $d \leq 3px$ are shown in green. From left to right: FisheyeSuperPoint, SuperPoint \cite{detone2018superpoint}, SIFT \cite{lowe2004}, ORB \cite{rublee2011orb}.}
\label{fig_4}
\end{figure*}

For the detection repeatability on the HPatches benchmark, FisheyeSuperPoint outperforms the classical detectors on the scenes with illumination changes. For the scenes with viewpoint changes, it outperforms the other detectors when a higher NMS is applied. FisheyeSuperPoint and SuperPoint consistently outperform FAST and Shi. While the repeatability values for all detectors are lower in \cite{detone2018superpoint} the ranking of the detectors is consistent. The results match the values reported by \cite{pautrat2021superpoint}.

The results of the homography and matching correctness on HPatches and RobotCar are shown in Table \ref{tab2}. In addition to FisheyeSuperPoint and SuperPoint, we compare the performance to SIFT \cite{lowe2004} and ORB \cite{rublee2011orb} which are implemented using OpenCV. NMS $= 8$ and $\epsilon = 3$ is applied for all experiments. The number of detected points is set to $k = 1000$ for HPatches and $k = 300$ for RobotCar. The HPatches images are resized to $480 \times 640$ and the RobotCar images to $512 \times 512$. 

Both FisheyeSuperPoint and SuperPoint show a superior performance compared to SIFT and ORB when used for descriptor matching on the RobotCar test data. This is particularly evident when taking into account the RMSE results as per \ref{eqn:rmse}, which indicate a high number of outliers for SIFT and Orb with $RMSE > 100$. The matching performance on RobotCar is also shown in Figure \ref{fig_4}, where matches with a euclidean distance of $d \leq 3px$ are indicated with green lines. SIFT outperforms the other algorithms for the homography correctness on HPatches. FisheyeSuperPoint ranks second and is superior to SuperPoint and ORB.

\section{\uppercase{Conclusion}}

This work describes the new FisheyeSuperPoint keypoint detection and description network which uses a pipeline to train and evaluate it directly on fisheye image datasets.
To enable the self-supervised training on fisheye images, fisheye warping is utilised. The fisheye image is mapped to a new, warped fisheye image through the intermediate step of projection to a unit sphere, with the camera's virtual pose being varied in six degrees of freedom. This process is embedded in an existing SuperPoint implementation \cite{pautrat2021superpoint} and trained on the RobotCar dataset \cite{maddern2017robotcar}. 

In order to compare the performance of FisheyeSuperPoint to other detectors, we introduce a method to evaluate keypoint detection repeatability and matching correctness on fisheye images. FisheyeSuperPoint consistently outperforms SuperPoint for the experiments on standard images (HPatches), especially in terms of homography correctness. This might be due to more variations in the RobotCar training data. Both FisheyeSuperPoint and SuperPoint perform similarly in our fisheye evaluations. This was unexpected and hints towards the robustness of the SuperPoint network, suggesting that it could be used for keypoint detection and description on fisheye images directly. Further evaluations on non-artificial data for descriptor matching correctness could provide a better insight into the performance of both networks.

While Harris and SIFT achieve higher repeatabilities and homography correctness than FisheyeSuperPoint in a few cases, our method comes with several advantages. One of those advantages is the adaptability of the network structure, which could be enhance with alternatives such as deformable convolutional layers \cite{dai2017defnets}. The adaptive manner of deformable convolutional layers could help increase detection and description performance under the influence of the radial distortion in fisheye images. 

Another opportunity is to incorporate FisheyeSuperPoint into multi-task visual perception networks like Omnidet \cite{kumar2021omnidet}. Multi-task networks can present advantages in computational complexity and performance by sharing base layers of a network, which will be enhanced with FisheyeSuperPoint. 

\section*{\uppercase{Acknowledgements}}

This publication has emanated from research conducted with the financial support of Science Foundation Ireland (SFI) under Grant number 18/CRT/6049 and 16/RI/3399. The opinions, findings and conclusions or recommendations expressed in this material are those of the author(s) and do not necessarily reflect the views of the Science Foundation Ireland.

\bibliographystyle{apalike}
{\small
\bibliography{main}}

\end{document}